%% file: main.tex
\documentclass[runningheads]{llncs}
 
\usepackage{eccv} 
  
\usepackage{eccvabbrv}

\usepackage{graphicx}
\usepackage{booktabs}

\usepackage[accsupp]{axessibility}  

\usepackage[pagebackref,breaklinks,colorlinks,citecolor=eccvblue]{hyperref}

\usepackage{orcidlink}

\usepackage{bm}

\usepackage{graphicx}
\usepackage{amsmath}
\usepackage{amssymb}
\usepackage{booktabs}
  
\usepackage{caption}
\usepackage{multirow}

\usepackage{wrapfig}

\usepackage{epsfig}
\usepackage{graphicx}
\usepackage{amsmath}
\usepackage{amssymb}
\usepackage{makecell}
\usepackage{caption}
\usepackage{wrapfig}
\usepackage{booktabs}
\usepackage[section]{placeins}

\input{preamble}
\begin{document} 
  
\title{HumANDiff: Articulated Noise Diffusion for Motion-Consistent Human Video Generation} 

\titlerunning{HumANDiff: Articulated Noise Diffusion}


\author{Tao Hu \and Varun Jampani}
 


\institute{Stability AI}

\maketitle

\newcommand{\nickname}{HumANDiff}

\input{pack/tables}
\input{pack/figures}
\input{pack/figures_sup}

\input{sec/abstract}
\input{sec/1_introduction}
\input{sec/related}
\input{sec/3_method}
\input{sec/4_results}
\input{sec/discussion}

%
%

\clearpage
\newpage

\input{pack/tables_sup}

\input{sec/sup_opt}

\input{sec/sup_exp_org}

\bibliographystyle{splncs04}
\bibliography{main_short}
\end{document}

%% file: preamble.tex
\usepackage{overpic}
\usepackage{enumitem} 
\usepackage{overpic} 
\usepackage{color}

\definecolor{turquoise}{cmyk}{0.65,0,0.1,0.3}
\definecolor{purple}{rgb}{0.65,0,0.65}
\definecolor{dark_green}{rgb}{0, 0.5, 0}
\definecolor{orange}{rgb}{0.8, 0.6, 0.2}
\definecolor{red}{rgb}{0.8, 0.2, 0.2}
\definecolor{darkred}{rgb}{0.6, 0.1, 0.05}
\definecolor{blueish}{rgb}{0.0, 0.3, .6}
\definecolor{light_gray}{rgb}{0.7, 0.7, .7}
\definecolor{pink}{rgb}{1, 0, 1}
\definecolor{greyblue}{rgb}{0.25, 0.25, 1}

\newcommand\cnum[1]{\raisebox{.5pt}{\textcircled{\raisebox{-0.9pt}{#1}}}}

\usepackage{blindtext}

\renewcommand{\paragraph}[1]{\vspace{1em}\noindent\textbf{#1}.}

%% file: pack/tables.tex
\newcommand{\tabmain}{
\begin{table*}[t]
\centering
\caption{Quantitative metrics for fashion video generation. "Dreampose*" denotes the result after sample finetuning. Note that pixel-level metrics (\eg, SSIM and L1) are more sensitive to human pose estimation accuracy, whereas perceptual metrics (\eg, LPIPS, FID, FVD) provide a more reliable evaluation for human generation tasks \cite{neuralactor}.}
\begin{tabular}{lcccccc}
\hline
 & {L1} $\downarrow$ & {SSIM} $\uparrow$ & {LPIPS} $\downarrow$ & {FID} $\downarrow$ & {FVD} $\downarrow$ & {FID-VID} $\downarrow$ \\
\hline \hline
MRAA~\cite{mraa}             & .0857 & .749 & .212 & 23.42 & 253.65 & -- \\
TPSMM~\cite{tpsmm}           & .0858 & .746 & .212 & 22.87 & 247.55 & -- \\
PIDM~\cite{pidm}             & .1098 & .713 & .288 & 30.28 & 238.75 & -- \\
BDMM~\cite{bdmm}             & .9180 & --   & .048 & --    & 148.30 & -- \\
DreamPose~\cite{dreampose}   &  --   & .8790 & .111 & --    & 279.60 & -- \\
DreamPose*~\cite{dreampose}  & .0256 & .8850 & .068 & 13.04    & 238.75    & -- \\
SVD-I2V~\cite{svd}           & -- & .8940   & .095 & - & 175.4 & -- \\
CogVideoX5B-I2V~\cite{cogvideox} 
& .0979 & .827 & .133 & 15.59 & 397.06 & 8.57 \\
Go-With-The-Flow~\cite{goflow}
& .1919 & .756 & .241 & 15.56 & 228.23 & 15.77 \\
\textbf{Ours} 
& \textbf{.0364} & \textbf{.906} & \textbf{.057} & \textbf{7.94} & \textbf{84.10} & \textbf{4.68} \\
\hline
\end{tabular}
\vspace{-0.16in}
\label{tab:main}
\end{table*}
}

\newcommand{\tabAb}{
\begin{table*}[h]
\small
\centering
\caption{Ablation study results for JAML and GMCL. Note that perceptual metrics (\eg, LPIPS, FID, FVD) are generally more reliable than pixel-level metrics (\eg, SSIM and L1) for evaluating human generation tasks \cite{neuralactor}.}
\begin{tabular}{lcccccc}
\hline
& {L1} $\downarrow$ & {SSIM} $\uparrow$ & {LPIPS} $\downarrow$ & {FID} $\downarrow$ & {FVD} $\downarrow$ & {FID-VID} $\downarrow$ \\
\hline \hline 
Base + Articulated Noise Sampling 
& .0378 & .903 & .071 & 10.47 & 12.03 & 131.30 \\

+ JAML 
& .0371 & .905 & .065 & 9.02 & 10.00 & 117.68 \\

+ GMCL 
& \textbf{.0364} & \textbf{.906} & \textbf{.057} & \textbf{7.94} & \textbf{4.68} & \textbf{84.10} \\
\hline
\end{tabular}
\label{tab:ab}
\end{table*}
}

%% file: pack/figures.tex
\newcommand{\figTeaser}{
\begin{figure}[t]
	\begin{center}
		\includegraphics[width=.7\linewidth]{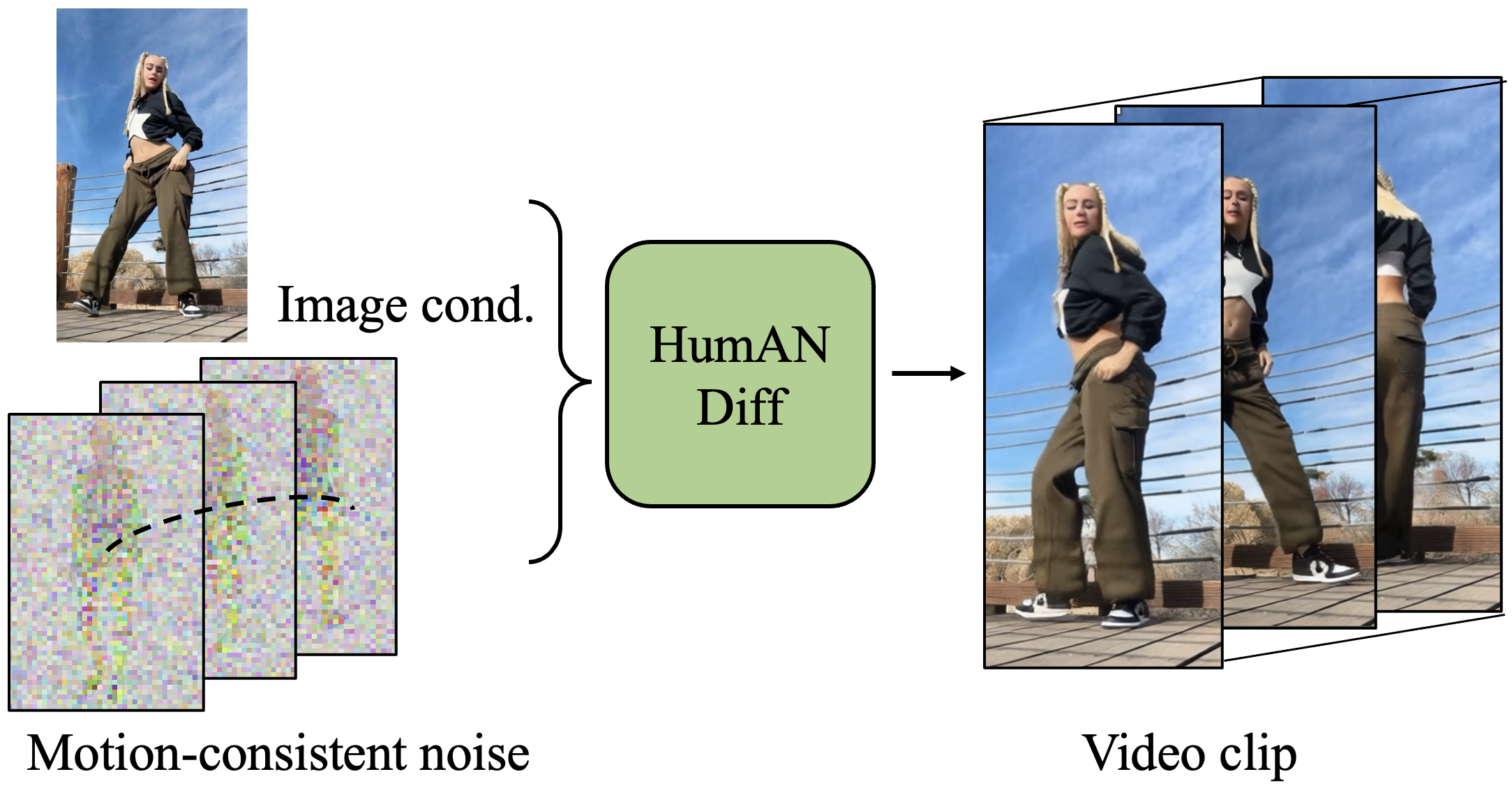}
	\end{center}
	\vspace{-0.13in}
	\caption{HumANDiff generates temporally consistent human videos from a single image by sampling motion-consistent noise on the UV manifold of a parametric body model (\eg, SMPL \cite{smpl}), enabling intrinsic motion control without explicit motion adapters.} 
	\label{fig:teaser}
	\vspace{-0.20in}
\end{figure}
}

\newcommand{\figFramework}{
\begin{figure*}[t]
	\begin{center}
		\includegraphics[width=\linewidth]{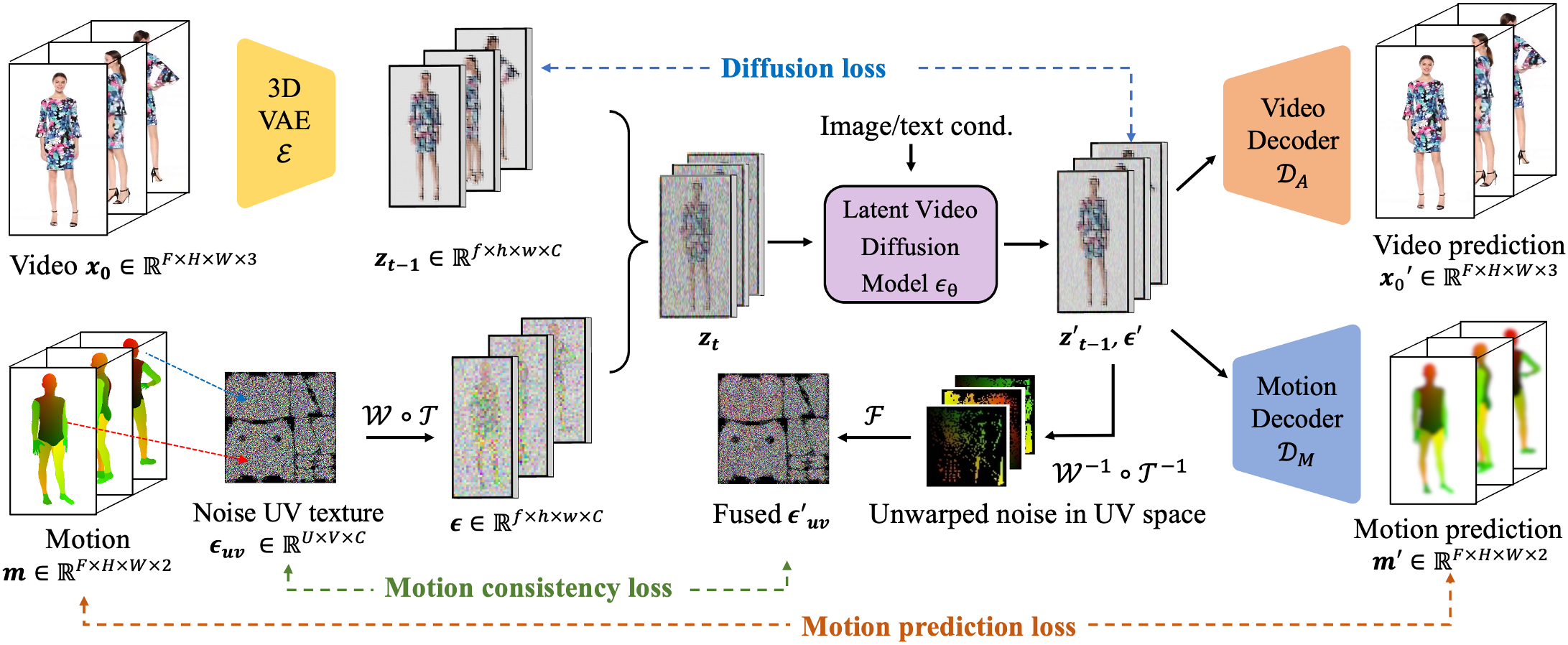}
	\end{center}
	\vspace{-0.14in}
	\caption{Training framework. During training, \nickname{} learns to generate motion-controllable human videos using a latent video diffusion model with four components:  \textbf{1) Video Encoding}: we employ a 3D VAE to encode video segments $\bm{x_0}$ into latent space. \textbf{2) Articulated Motion Consistent Noise Sampling} (Sec. \ref{sec:method_sample}): We model motions on the surface UV manifold of SMPL \cite{smpl} and integrate motion control $\bm{m_0}$ as a warping field to sample noise $\bm{\epsilon}$ from a noise texture UV map. \textbf{3) Joint Appearance-Motion Learning} (Sec. \ref{sec:method_ft}): The diffusion model is enhanced to learn both video and motion for latent denoising to improve motion coherence. \textbf{4) Geometric Motion Consistency Learning} (Sec. \ref{sec:method_ft}): We further propose a motion consistency loss to enforce motion consistency across frames between the predicted noise $\bm{\epsilon'_{uv}}$ and sampled noise $\bm{\epsilon_{uv}}$ in UV space.}
    \label{fig:framework}
	\vspace{-0.12in}
\end{figure*}
}

\newcommand{\figCmpMotionMain}{
\begin{figure}[t]
	\begin{center}
		\includegraphics[width=.85\linewidth]{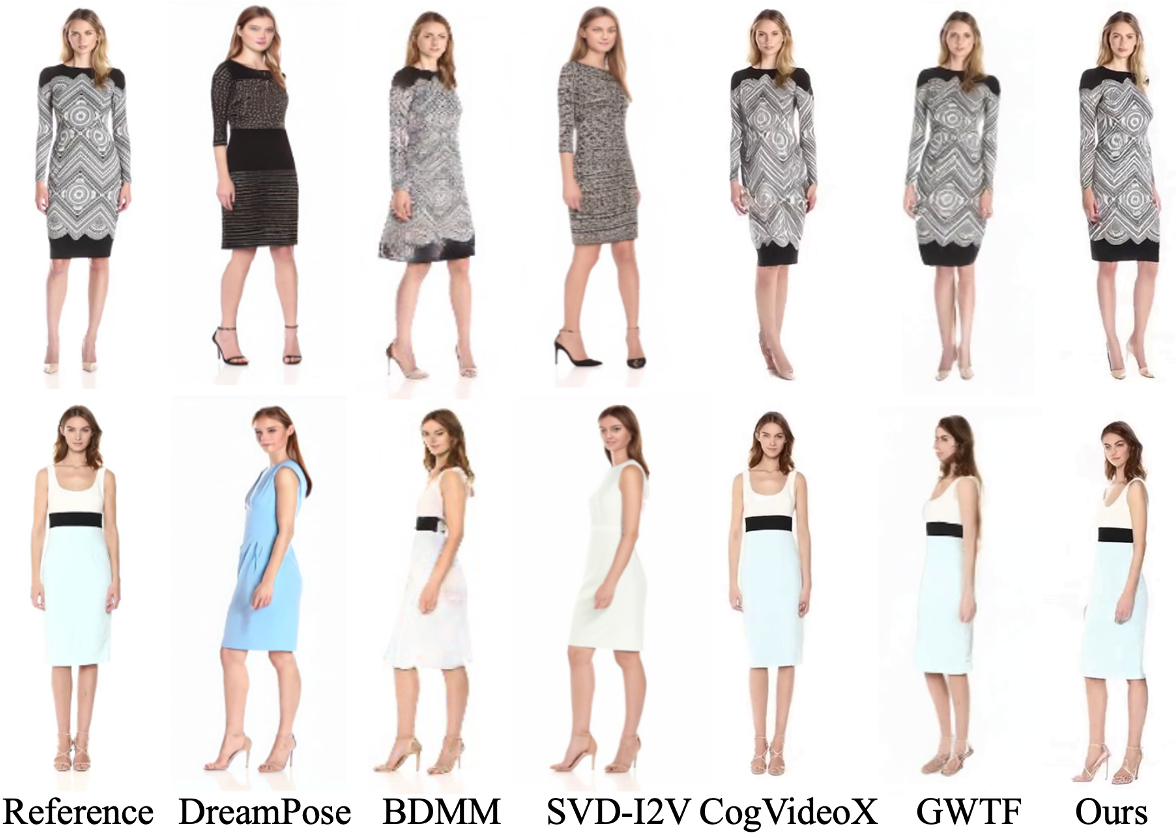}
	\end{center}
	\vspace{-0.14in}
	\caption{Qualitative comparisons for fashion video synthesis. We show the results of each methods under similar novel poses. Our method can faithfully preserve the appearance details of the reference image, \eg, face identity and clothing details.}
	\label{fig:cmp_main}
\end{figure}
}

\newcommand{\figAbMd}{
\begin{figure}[ht]
	\begin{center}
		\includegraphics[width=.8\linewidth]{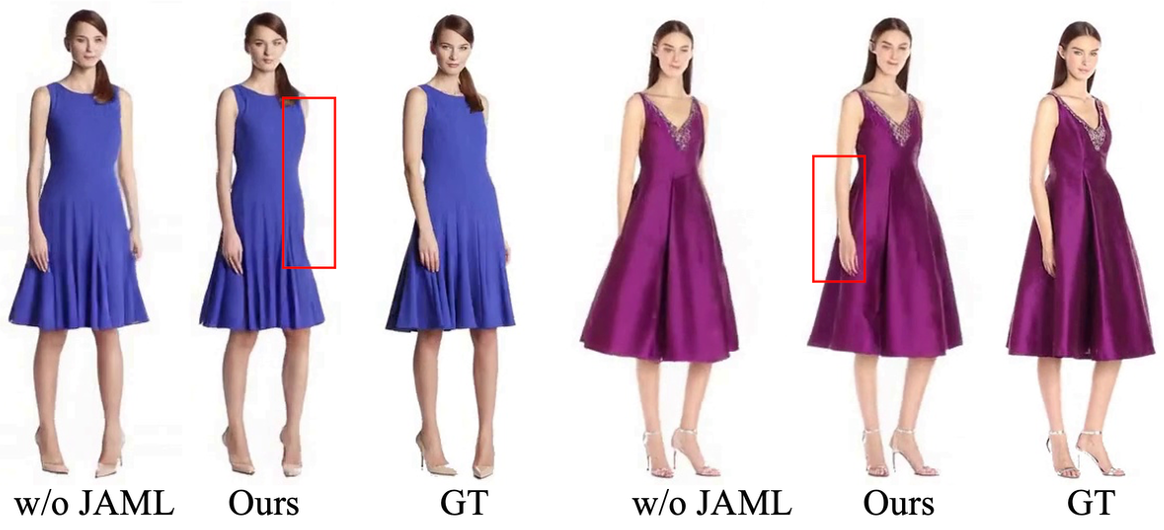}
	\end{center}
	\vspace{-0.12in}
\caption{Ablation study of Joint Appearance-Motion Learning (JAML). JAML improves motion coherence for faithful motion control, while the variant without JAML failed to generate accurate animation videos.}
\label{fig:ab_md}
	\vspace{-0.12in}
\end{figure}
}

\newcommand{\figAbMc}{
\begin{figure}[ht]
	\begin{center}
		\includegraphics[width=.75\linewidth]{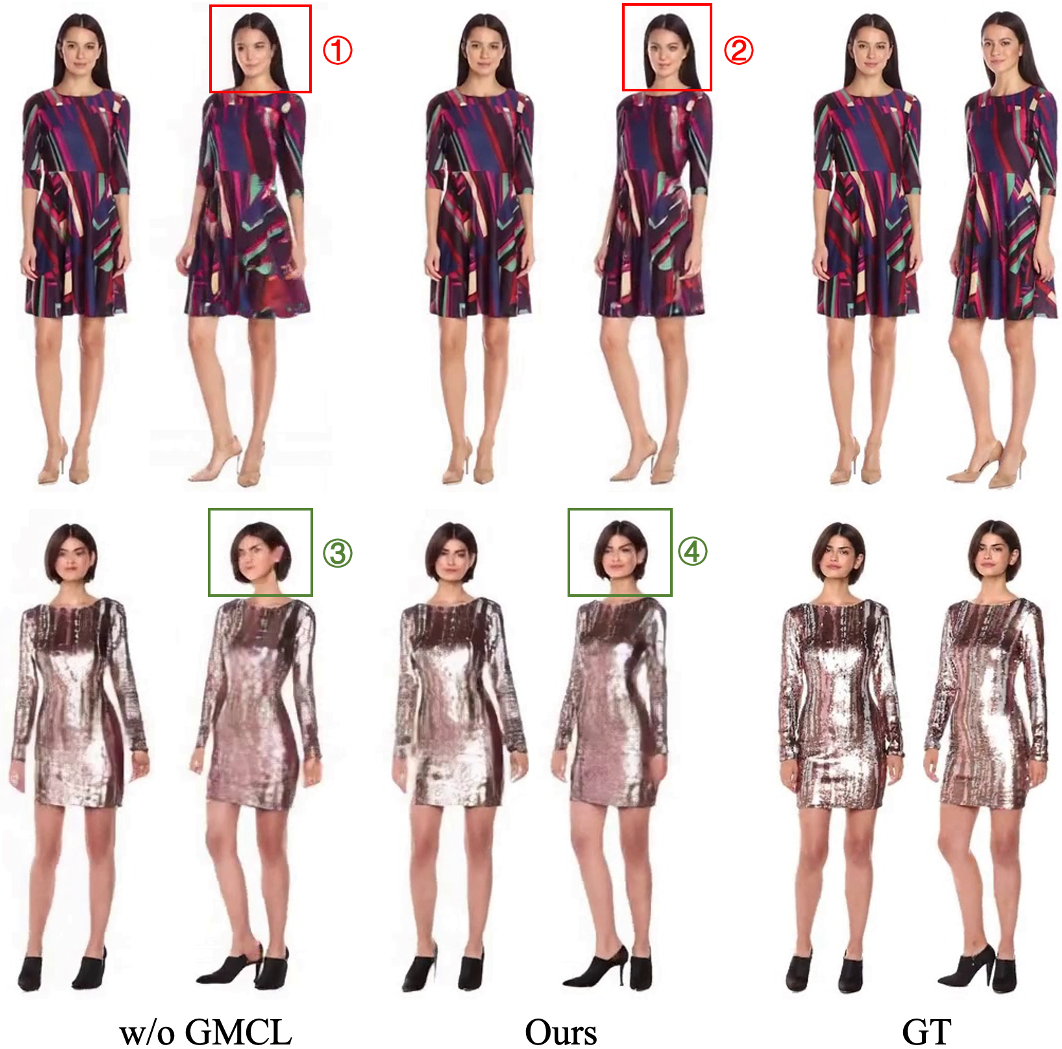}
	\end{center}
	\vspace{-0.12in}
\caption{Ablation study of Geometric Motion Consistency Learning (GMCL), where two frames at different time steps are shown. GMCL improves appearance consistency, \textit{i.e.}, it preserves the face identity details better (\cnum{2} vs. \cnum{1}, and \cnum{4} vs. \cnum{3}).}
\label{fig:ab_mc}
\end{figure}
}

\newcommand{\figSupCmpAAMotionFlow}{
\begin{figure*}[ht]
	\begin{center}
		\begin{tabular}{cc}
			\begin{minipage}[t]{\linewidth}
				\centering
				\includegraphics[width=.8\linewidth]{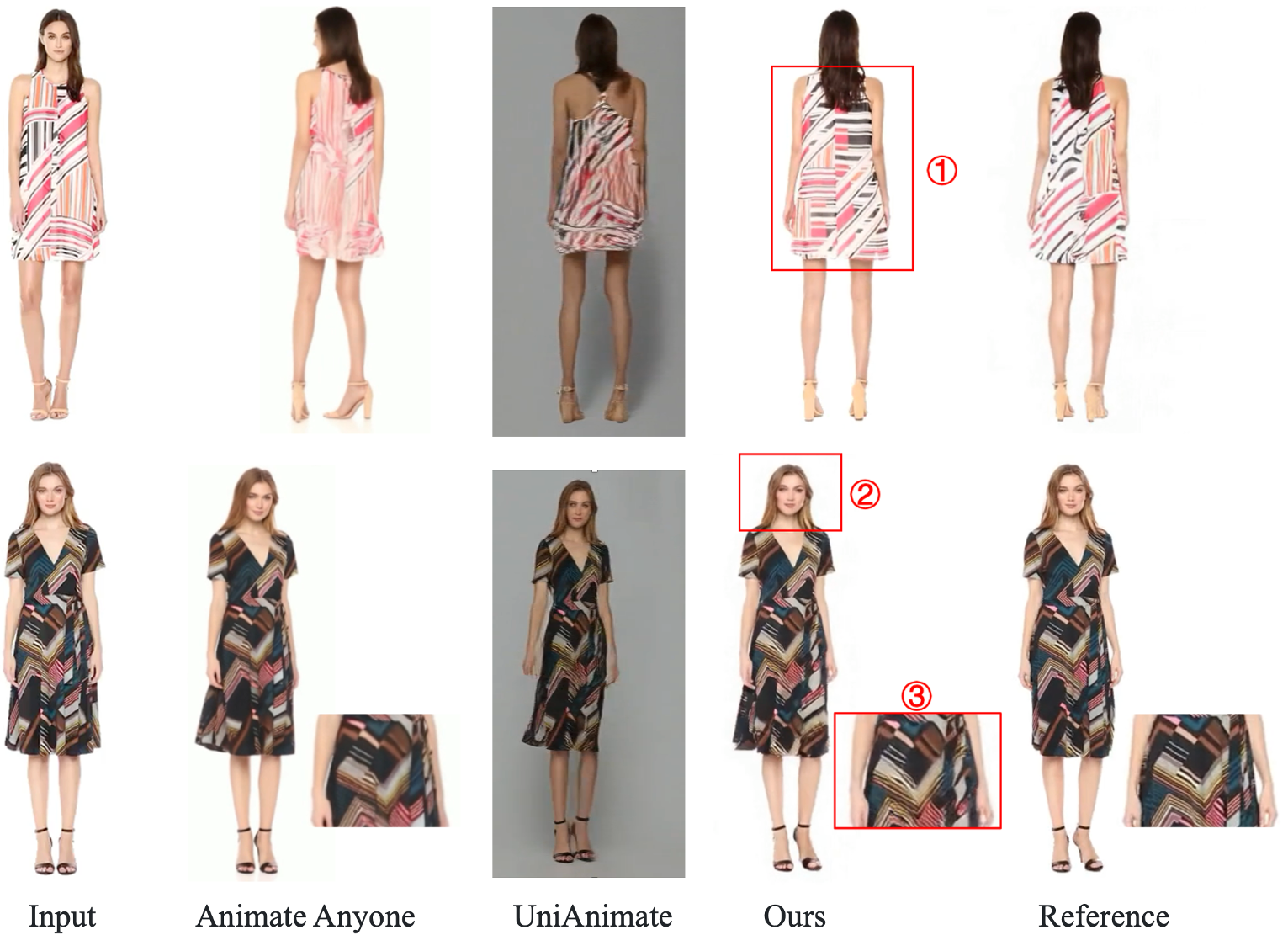}
				\caption{Comparisons against Animate Anyone \cite{hu2024animate} and Wan2.1 based UniAnimate \cite{wang2025unianimate}. Our method can generate high-quality clothing details (\cnum{1}, \cnum{3}), and face details (\cnum{2}).}
				\label{fig:sup_cmpaa}
			\end{minipage}
			
			\\ \\[0.2em]

			\begin{minipage}[t]{\linewidth}
				\centering
				\includegraphics[width=.65\linewidth]{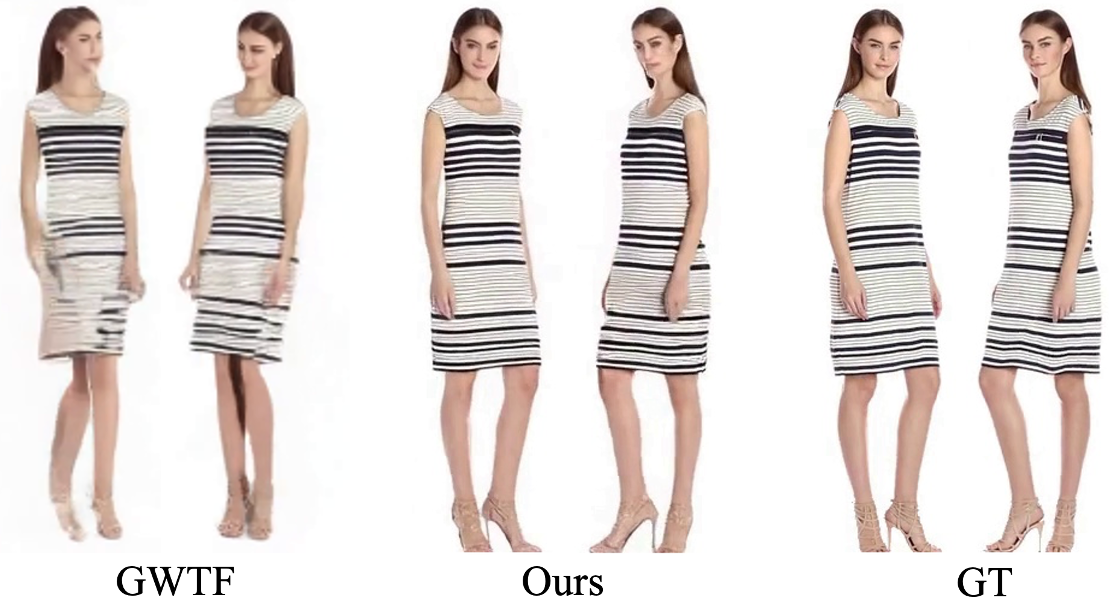}
				\caption{Qualitative comparisons for appearance consistency of different noise warping methods, where two different frames are visualized for evaluations. Compared with 2D optical flow based noise warping GWTF \cite{goflow}, our 3D articulated noise warping enables more precise motion control, and generates more consistent rendering results for complex  textures.}
				\label{fig:cmp_goflow}
			\end{minipage}
		\end{tabular}
	\end{center}
		\vspace{-0.4in}
\end{figure*}
}

\newcommand{\figCmpMotionMainAndTabMain}{
\begin{figure*}[ht]
	\centering
	\begin{minipage}[t]{\linewidth}
		\centering
		\includegraphics[width=.8\linewidth]{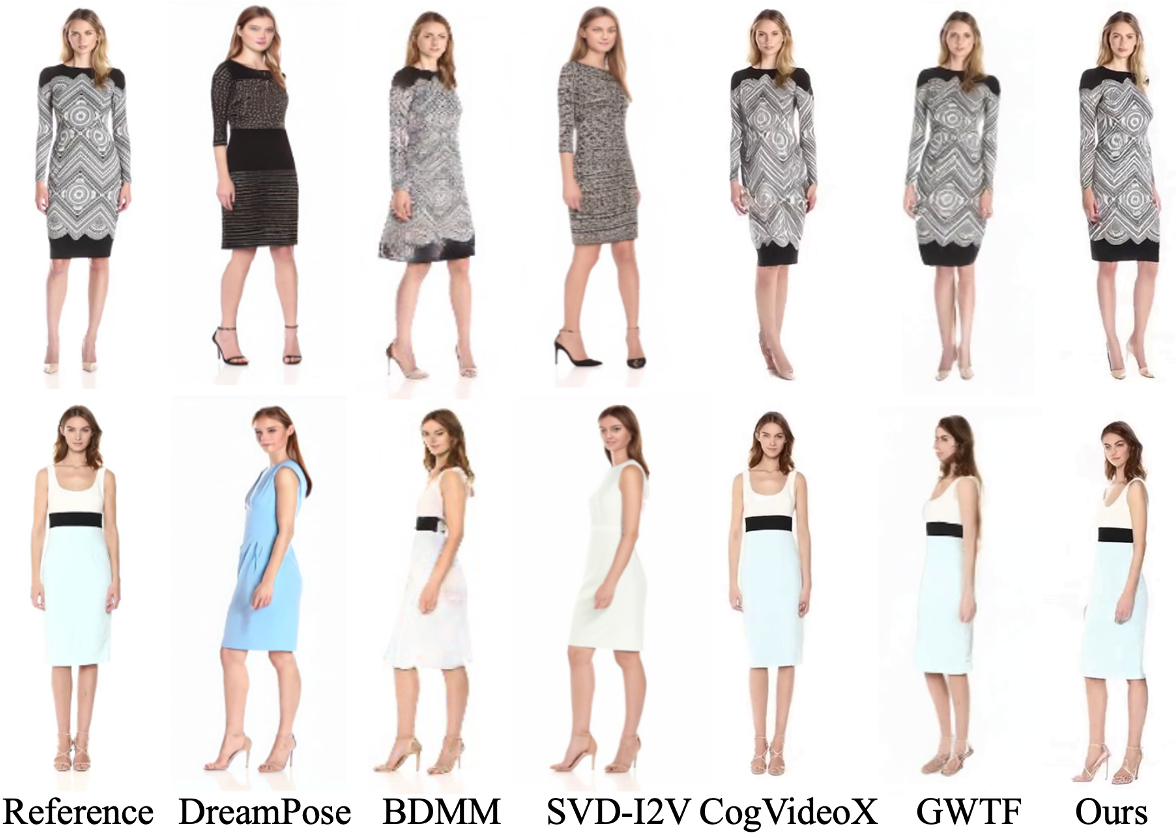}
		\captionof{figure}{Qualitative comparisons for fashion video synthesis. We show the results of each methods under similar novel poses. Our method can faithfully preserve the appearance details of the reference image, \eg, face identity and clothing details.}
		\label{fig:cmp_main}
	\end{minipage}

	\vspace{0.2em}

	\begin{minipage}[t]{\linewidth}
		\centering
		\captionof{table}{Quantitative metrics for fashion video generation. ``Dreampose*'' denotes the result after sample finetuning. Note that pixel-level metrics (\eg, SSIM and L1) are more sensitive to human pose estimation accuracy, whereas perceptual metrics (\eg, LPIPS, FID, FVD) provide a more reliable evaluation for human generation tasks \cite{neuralactor}.}
		\label{tab:main}
		\begin{tabular}{lcccccc}
		\hline
		 & {L1} $\downarrow$ & {SSIM} $\uparrow$ & {LPIPS} $\downarrow$ & {FID} $\downarrow$ & {FVD} $\downarrow$ & {FID-VID} $\downarrow$ \\
		\hline \hline
		MRAA~\cite{mraa}             & .0857 & .749 & .212 & 23.42 & 253.65 & -- \\
		TPSMM~\cite{tpsmm}           & .0858 & .746 & .212 & 22.87 & 247.55 & -- \\
		PIDM~\cite{pidm}             & .1098 & .713 & .288 & 30.28 & 238.75 & -- \\
		BDMM~\cite{bdmm}             & .9180 & --   & .048 & --    & 148.30 & -- \\
		DreamPose~\cite{dreampose}   &  --   & .8790 & .111 & --    & 279.60 & -- \\
		DreamPose*~\cite{dreampose}  & .0256 & .8850 & .068 & 13.04    & 238.75    & -- \\
		SVD-I2V~\cite{svd}           & -- & .8940   & .095 & - & 175.4 & -- \\
		CogVideoX5B-I2V~\cite{cogvideox}
		& .0979 & .827 & .133 & 15.59 & 397.06 & 8.57 \\
		Go-With-The-Flow~\cite{goflow}
		& .1919 & .756 & .241 & 15.56 & 228.23 & 15.77 \\
		\textbf{Ours}
		& \textbf{.0364} & \textbf{.906} & \textbf{.057} & \textbf{7.94} & \textbf{84.10} & \textbf{4.68} \\
		\hline
		\end{tabular}
	\end{minipage}
	\vspace{-0.2in}
\end{figure*}
}

%% file: pack/figures_sup.tex
\newcommand{\figSupDf}{
	\vspace{-0.14in}
\begin{figure*}[th]
		\vspace{-0.14in}
	\begin{center}
		\includegraphics[width=.8\linewidth]{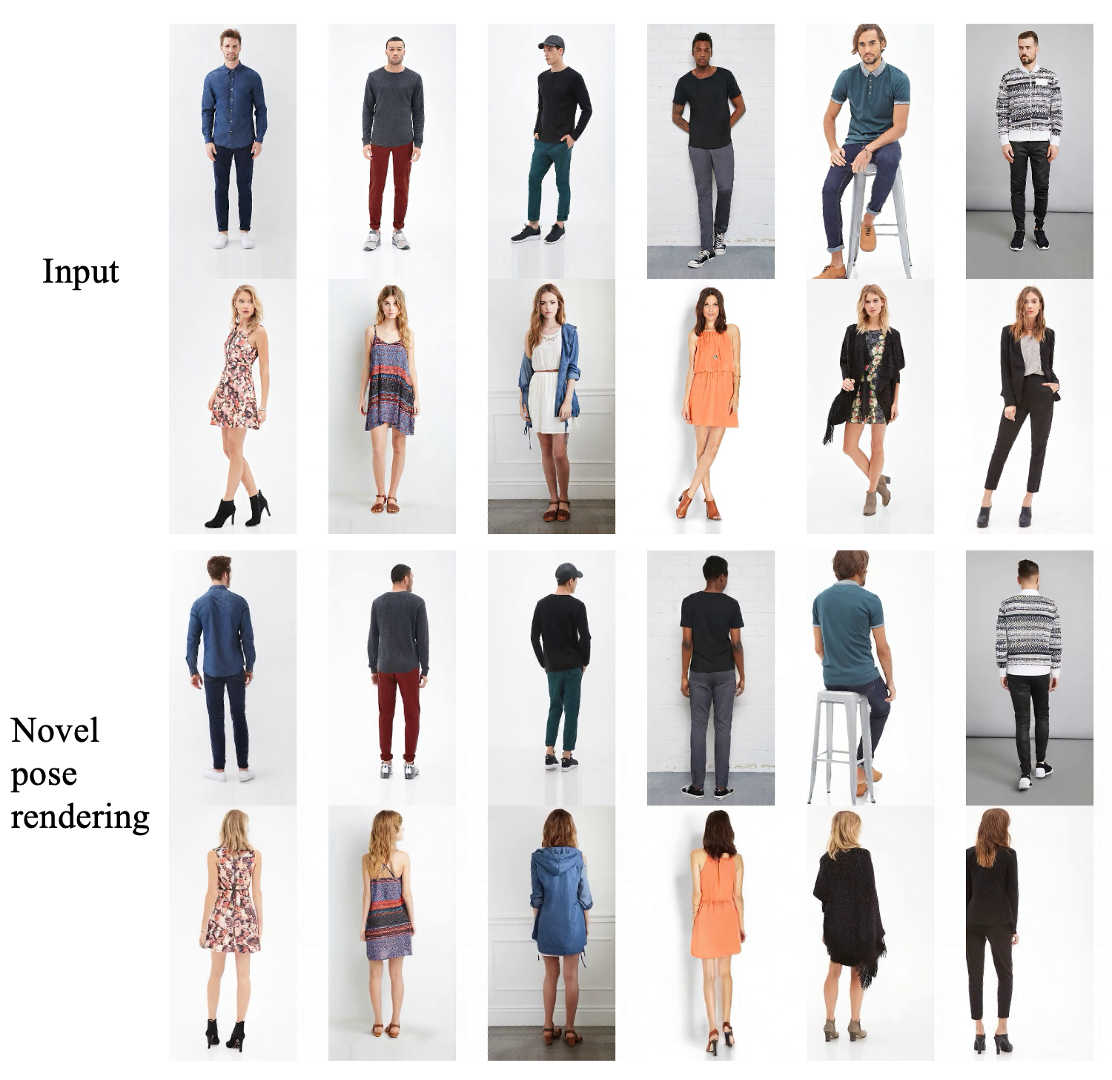}
	\end{center} 
	\vspace{-0.18in}
	\caption{Generalization on out-of-domain data (DeepFashion  \cite{Liu2016DeepFashion}). Our method trained on fashion videos generalizes well to unseen out-of-domain data, and  synthesizes human novel pose images from a single image of humans in diverse clothing styles.}
	\label{fig:sup_df}
\end{figure*}
}

\newcommand{\figSupSm}{
\begin{figure*}[h]
	\begin{center}
		\includegraphics[width=\linewidth]{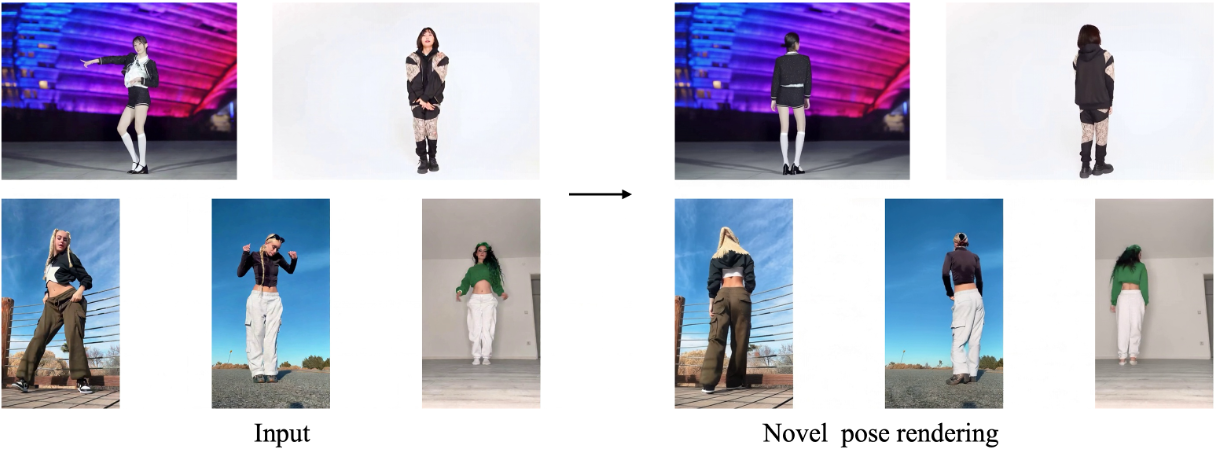}
	\end{center}
    \vspace{-0.18in}
	\caption{Generalization on in-the-wild social media videos. Our method trained on fashion videos generalizes well to unseen out-of-domain social media videos, and synthesizes novel pose images in diverse backgrounds.)}
	\label{fig:sup_sm}
\end{figure*}
}

\newcommand{\figGeneration}{
\begin{figure}[ht]
    \centering
    \begin{subfigure}{\textwidth}
        \centering
        \includegraphics[width=.75\linewidth]{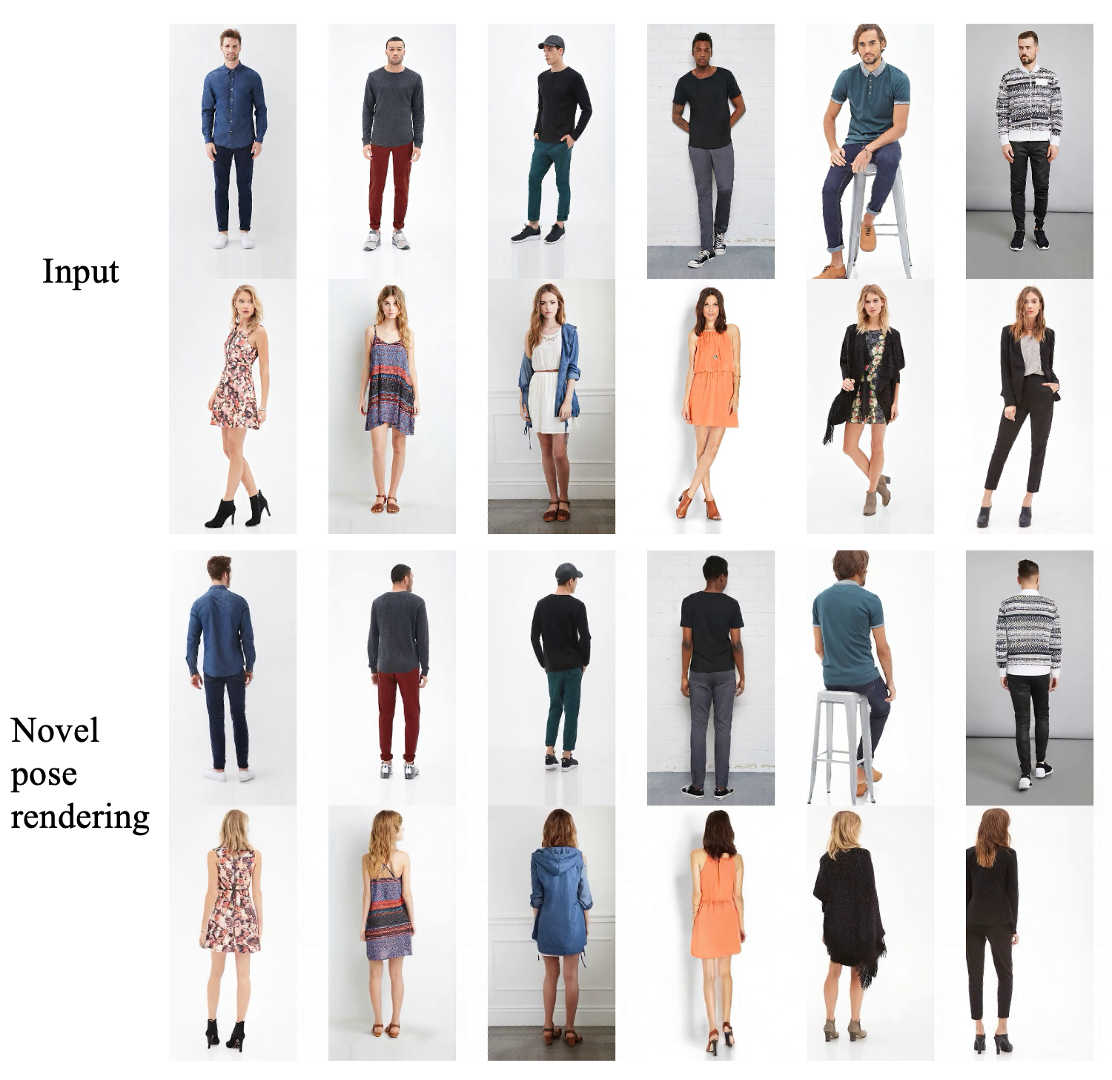}
        \caption{Generalization on out-of-domain data (DeepFashion \cite{Liu2016DeepFashion}).}
        \label{fig:sup_df}
    \end{subfigure}
    \\[0.5em]
    \begin{subfigure}{\textwidth}
        \centering
        \includegraphics[width=\linewidth]{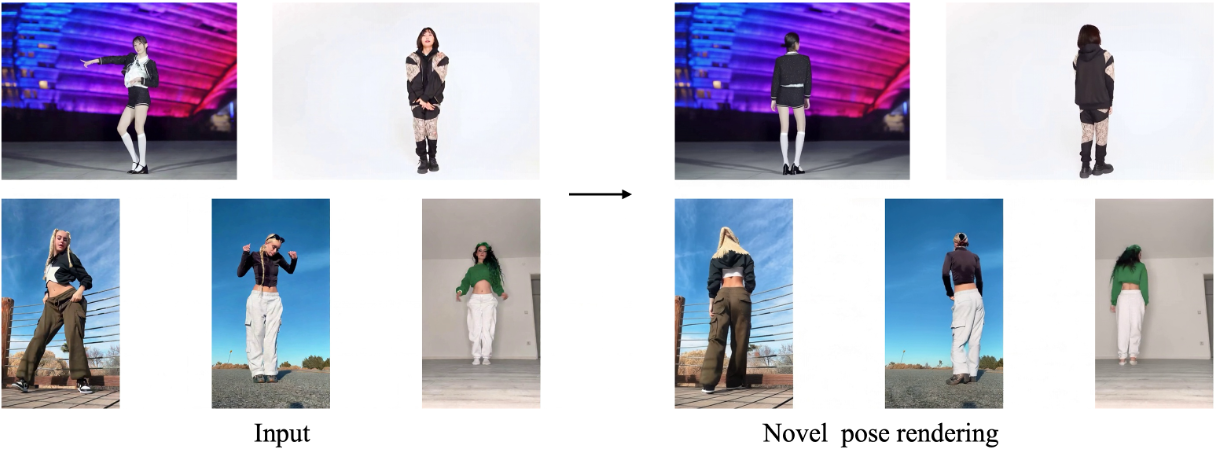}
        \caption{Generalization on in-the-wild videos.}
        \label{fig:sup_sm}
    \end{subfigure}
    \vspace{-0.16in}
    \caption{Generalization results. Our method trained on fashion videos generalizes well to unseen out-of-domain data, and synthesizes human novel pose videos from a single image  in diverse clothing styles (a), and diverse in-the-wild backgrounds (b).}
    \label{fig:sup_generation}
		\vspace{-0.2in}
\end{figure}
}

\newcommand{\figSupCmpAA}{
\begin{figure*}[t]
	\begin{center}
		\includegraphics[width=.85\linewidth]{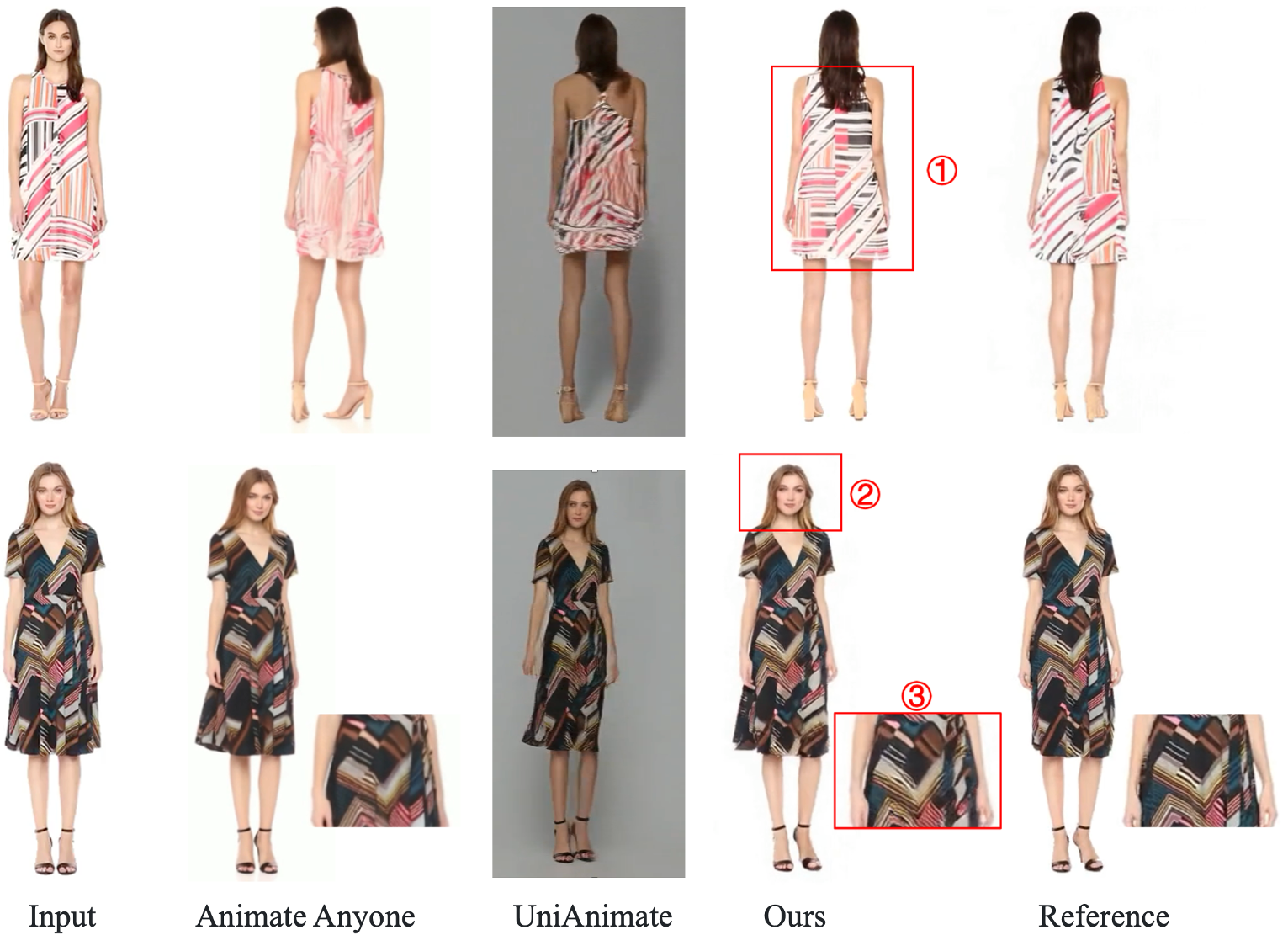}
	\end{center}
    \vspace{-0.16in}
	\caption{Comparisons against Animate Anyone \cite{hu2024animate} and Wan2.1 based UniAnimate \cite{wang2025unianimate}. Our method can generate high-quality clothing details (\cnum{1}, \cnum{3}), and face datails (\cnum{2}).}
	\label{fig:sup_cmpaa}
\end{figure*}
}

\newcommand{\figSupCmpAAOrg}{
\begin{figure*}[t]
	\begin{center}
		\includegraphics[width=.6\linewidth]{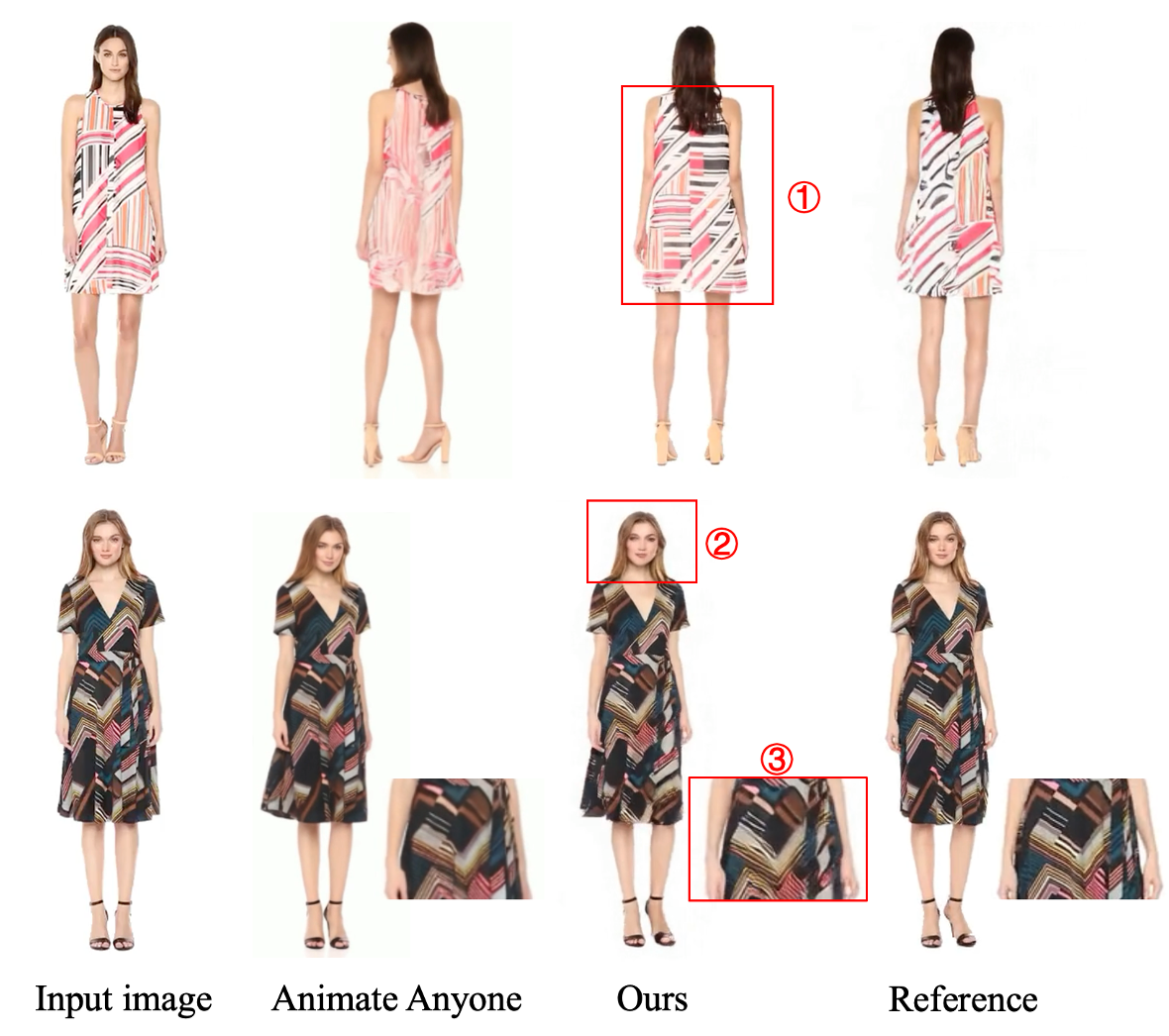}
	\end{center}
	\caption{Comparisons against Animate Anyone \cite{hu2024animate}. Our method can generate high-quality clothing details (\cnum{1}, \cnum{3}), and face datails (\cnum{2}).}
	\label{fig:sup_cmpaa}
\end{figure*}
}

\newcommand{\figSupAbDeg}{
\begin{figure*}[th]
	\begin{center}
		\includegraphics[width=.9\linewidth]{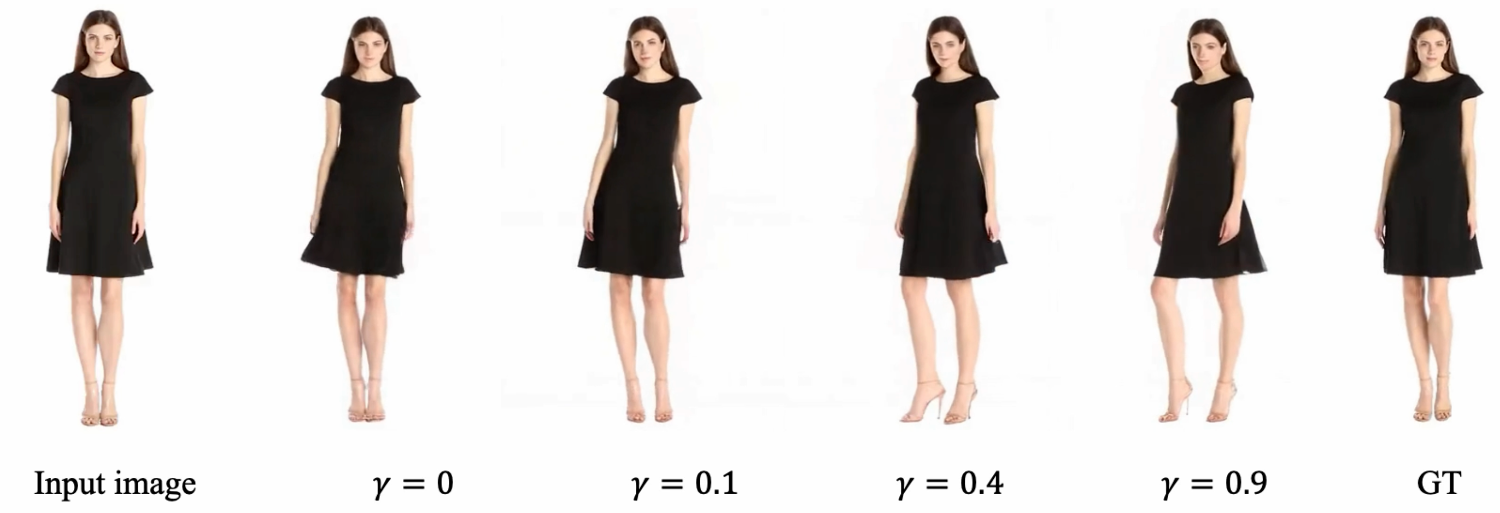}
	\end{center}
	\caption{Ablation study of noise degradation. We set $\gamma = 0.1$ to achieve a balance between motion control and video quality.}
	\label{fig:sup_ab_deg}
\end{figure*}
}

\newcommand{\figVAE}{
\begin{figure*}[h]
	\begin{center}
		\includegraphics[width=.6\linewidth]{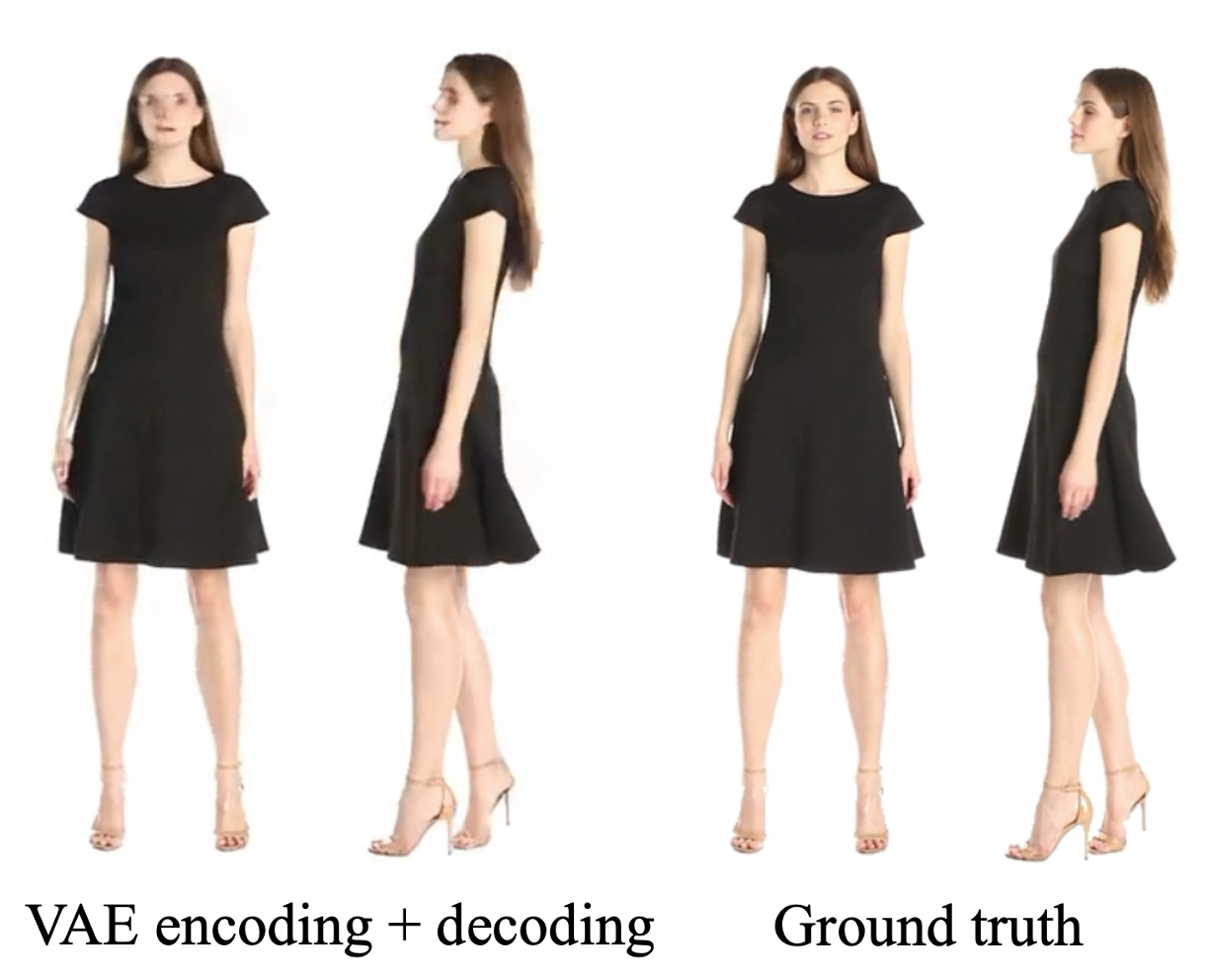}
	\end{center}
	\caption{Illustration of video VAE encoding in CogVideoX. Fine details such as faces and hands are often lost during video encoding, which makes it challenging for our model to generate high-quality face and hand details.}
	\label{fig:vae}
\end{figure*}
}

\newcommand{\figWanCog}{
\begin{figure*}[t]
	\begin{center}
		\includegraphics[width=.8\linewidth]{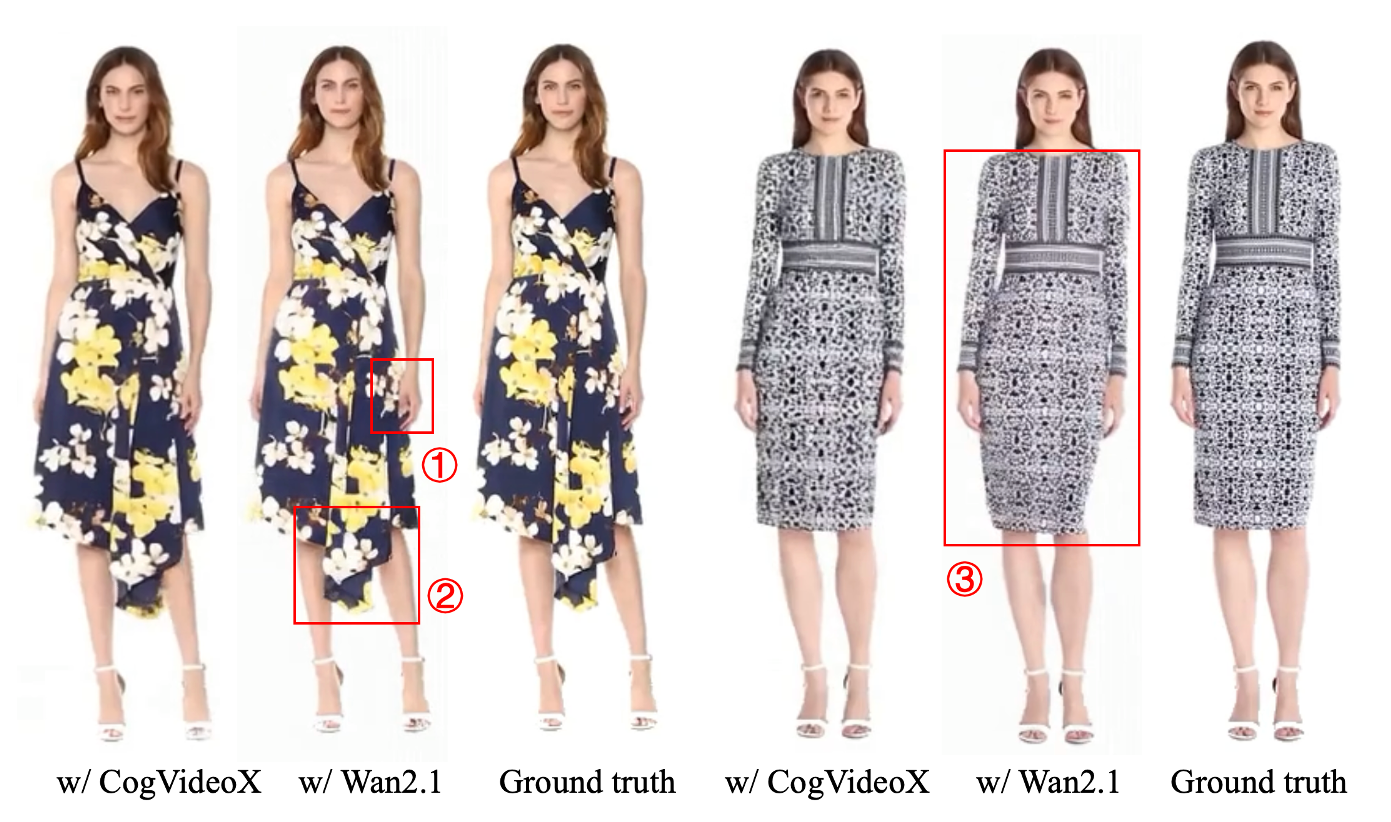}
	\end{center}
	\caption{Generation results of model variants trained on Wan2.1-I2V-14B and CogVideoX-I2V-5B. Our method is model agnostic, and the model trained on Wan2.1-I2V-14B generates higher quality appearance details (\cnum{1}\cnum{2}\cnum{3}).}
	\label{fig:wancog}
\end{figure*}
}

%% file: sec/abstract.tex
\begin{abstract}

Despite tremendous recent progress in human video generation, generative video diffusion models still struggle to capture the dynamics and physics of human motions faithfully. In this paper, we propose a new framework for human video generation, \textbf{\nickname}, which enhances the human motion control with three key designs: \textbf{1) Articulated motion-consistent noise sampling} that correlates the spatiotemporal distribution of latent noise and replaces the unstructured random Gaussian noise with 3D articulated noise sampled on the dense surface manifold of a statistical human body template. It inherits body topology priors for spatially and temporally consistent noise sampling. \textbf{2) Joint appearance-motion learning} that enhances the standard training objective of video diffusion models by jointly predicting pixel appearances and corresponding physical motions from the articulated noises. It enables high-fidelity human video synthesis, \eg, capturing motion-dependent clothing wrinkles. \textbf{3) Geometric motion consistency learning} that enforces physical motion consistency across frames via a novel geometric motion consistency loss defined in the articulated noise space. \nickname{} enables scalable controllable human video generation by fine-tuning video diffusion models with articulated noise sampling. Consequently, our method is agnostic to diffusion model design, and requires no modifications to the model architecture. During inference, HumANDiff enables image-to-video generation within a single framework, achieving intrinsic motion control without requiring additional motion modules. Extensive experiments demonstrate that our method achieves state-of-the-art performance in rendering motion-consistent, high-fidelity humans with diverse clothing styles. Project page: \href{https://taohuumd.github.io/projects/HumANDiff/}{taohuumd.github.io/projects/HumANDiff}.


\end{abstract}  

%% file: sec/1_introduction.tex
\section{Introduction}
\label{sec:intro}

\figTeaser  
  
Human video generation has garnered significant attention in recent years, with widespread applications in areas such as AR/VR, animation, video games, and film production. Recently, with the rapid development of diffusion models, especially latent diffusion models \cite{sd, svd, cogvideox}, video generation has achieved remarkable progress. However, generating realistic human motion videos is arguably challenging, as it requires not only high-quality appearance details but also smooth temporal consistency. Despite notable advances, existing methods for human video generation still struggle to capture the dynamics and physics of human motion accurately.

Within the domain of motion-controllable human video diffusion models, current approaches typically fall into three categories: (1) Learning a human-specific model from scratch \cite{shao2024human4dit,shao2024isa4d}, which generally requires a tremendous amount of training data and expensive computing resources. (2) Training a specific pose module to guide a pre-trained image diffusion model (e.g., Stable Diffusion \cite{sd}) for human video generation \cite{zhu2024champ,hu2024animate}. However, the image diffusion model cannot accurately capture the temporal motion consistency, and the extra introduced pose module reduces inference efficiency. (3) Leveraging generic video diffusion models with text prompts \cite{cogvideox} or motion-conditioned noise for fine-grained motion control \cite{goflow,videoeqm}, which preserves temporal consistency but struggles to faithfully capture the dynamics and physics of human motions due to the intricate human body structures. 

To address these limitations, we propose a new method \nickname{} that leverages a video diffusion prior for generalizable human video synthesis which avoids the limitations of expensive training on a large volume of data. In addition, \nickname{} requires no modifications to the base model architecture, \ie, it does not require training a specific pose module, which allows \nickname{} to fully leverage the video diffusion prior for intrinsic motion control and eliminates the need for additional pose processing during inference as shown in Fig. \ref{fig:teaser}. 
 
We achieve this by correlating the spatiotemporal distribution of latent noise, similar to the noise warping approaches \cite{goflow, videoeqm, howdo}. However, we illustrate that the existing warping approaches (\eg optical flow based warping \cite{goflow}) do not work for human motions due to the intricate human body structure and non-rigid motions. Instead, we enhance the video diffusion model to capture the dynamics of human motion with three key designs: \textbf{1)} Articulated motion-consistent noise sampling; \textbf{2)} Joint appearance-motion learning; and  \textbf{3)} Geometric motion consistency training. 

Firstly, in contrast to existing video diffusion methods that learn human motions from unstructured random noises ignoring the human body structure and temporal motion consistency, we propose to model motions in a structured unified space, \ie, the surface manifold of a 3D parametric human body mesh (\eg SMPL \cite{smpl}) which can be rendered as texture UV map in 2D space. We sample Gaussian noise on the texture noise UV map, which inherits articulated body topology priors for motion generalization and is shared across motion frames for temporal consistency. This allows us to replace the unstructured random Gaussian noise with structure-aware motion-conditioned and spatially and temporally consistent noise, making it easier for the diffusion model to learn motions.

Secondly, different from conventional pixel reconstruction objective that biases models toward appearance fidelity at the expense of motion coherence \cite{surmo,videojam}, we propose a joint appearance-motion learning strategy that encourages the video diffusion model to jointly predict pixel appearances and corresponding motions from the articulated noise space. In particular, in addition to the video decoder, we propose a motion decoder to predict physical motions from denoised latents. This physical motion learning enables high-fidelity human video synthesis, \eg, capturing motion-dependent clothing wrinkles for loose clothing such as dresses. We illustrate that the joint appearance-motion learning strategy can improve the motion coherence for faithful motion control.

Thirdly, we enhance the standard training objective that focuses on the pixel reconstruction quality while failing to capture the temporal motion consistency across frames, by proposing a geometric motion consistency training strategy. In particular, we propose a novel geometric motion consistency loss defined in the articulated noise space, which enforces motion consistency across frames while maintaining per-frame pixel quality. Experimental results demonstrate that the motion consistency supervision can significantly improve the temporal motion consistency by a large margin.
  
\nickname{} is built upon a video diffusion model, and it does not require any changes to the diffusion pipeline or architecture, and uses exactly the same amount of memory and runtime as the base model. This enables scalable training for controllable human video generation. 
 
In summary, our contributions are:

 
\textbf{1)} A novel model-agnoistic framework that extends base video diffusion models for controllable and longer human video generation without compromising efficiency, which is achieved by correlating the spatiotemporal distribution of latent noise on the surface manifold of a statistical human body template, which integrates motion control with articulated noise warping for spatially and temporally consistent noise modelling.

\textbf{2)} A joint appearance-motion learning strategy that improves the motion coherence by jointly predicting pixel appearances and corresponding physical motions from the articulated noises during training. 

\textbf{3)} A novel motion consistency training strategy that proposes a geometric motion consistency loss to enforce physical motion consistency while maintaining per-frame pixel quality.

\textbf{4)} A one-stop solution for image-to-video animation generation without requiring additional motion modules for intrinsic motion control at inference, and comprehensive experiments demonstrate the state-of-the-art results of our method for controllable human video generation.

%% file: sec/related.tex
\section{Related Work}

Our approach is closely related to several subfields of visual computing. Below, we discuss some of the most relevant connections.

\noindent {\bf Diffusion-Based Image and Video Synthesis.}
Diffusion models have achieved remarkable results in high-quality image~\cite{ho2020denoising, song2020score, rombach2022high, zhang2023adding, wang2024instantid} and video synthesis~\cite{lei2024comprehensive, shao2024human4dit, zhu2024champ, wang2023disco, icsik2023humanrf, zhang20254diffusion}. Latent diffusion models, such as Stable Diffusion~\cite{sd}, improve efficiency by operating in a compact latent space rather than directly on high-resolution pixels.

Video generation requires modeling both spatial fidelity and temporal consistency. Recent methods address this by adding temporal layers to pre-trained image diffusion models~\cite{singer2022make, guo2023animatediff, wu2023tune, wang2024magicvideo, blattmann2023stable} or adopting transformer-based architectures for sequential video modeling~\cite{yan2021videogpt, yu2023magvit}. For example, Stable Video Diffusion~(SVD)~\cite{blattmann2023stable}, CogVideoX~\cite{cogvideox} and Wan2.1 \cite{wan2025} extend latent diffusion to videos by integrating temporal modules and fine-tuning on large-scale video datasets, providing an open-source framework for high-fidelity video generation. CogVideoX~\cite{cogvideox} and Wan2.1 \cite{wan2025} adopt diffusion transformer architecture, and  encode videos into a compact latent space using a 3D video VAE and incorporates temporal modeling, enabling efficient and scalable video synthesis while preserving visual quality and temporal coherence. In this work, we extend CogVideoX and Wan2.1 for motion-guided human video generation from single images with fine-grained motion control.

\noindent {\bf Human Animation Video Generation.}
Human video generation aims to animate static images into realistic, temporally coherent sequences. Methods can be broadly categorized into 3D rendering and 2D image-based approaches.

3D rendering approaches~\cite{neuralbody, neuralactor, Peng2021AnimatableNR, narf, anerf, Chen2021AnimatableNR} combine geometry reconstruction with view synthesis using volume rendering, but these methods are computationally intensive. Hybrid 3D-GAN strategies~\cite{Niemeyer2021GIRAFFERS, Zhou2021CIPS3DA3, Gu2021StyleNeRFAS, OrEl2021StyleSDFH3, Hong2021HeadNeRFAR, Chan2021EfficientG3, hvtr, hvtrpp} improve efficiency by combining geometry-aware rendering with generative models. Gaussian Splatting approaches~\cite{qiu2024anigs, qiu2025lhm} allow real-time rendering of human avatars.

2D approaches use GANs to map poses—represented as skeletal renderings~\cite{edn, SiaroSLS2017, Pumarola_2018_CVPR, KratzHPV2017, zhu2019progressive, vid2vid}, dense meshes~\cite{Liu2019, vid2vid, liu2020NeuralHumanRendering, feanet, Neverova2018, Grigorev2019CoordinateBasedTI}, or joint heatmaps~\cite{MaSJSTV2017, Aberman2019DeepVP, Ma18}—to realistic images. Temporal stability can be improved by incorporating SMPL priors~\cite{smpl, dnr, smplpix, egorend, anr}. However, these methods do not explicitly model 3D geometry and struggle with self-occlusions.

Diffusion-based human video generation methods, such as Human4DiT~\cite{shao2024human4dit} and ISA4D~\cite{shao2024isa4d}, achieve higher quality than GAN-based approaches but often require massive training data and computational resources. To reduce these requirements, some methods train pose modules to guide pre-trained image diffusion models (e.g., Stable Diffusion~\cite{sd}). These methods use either 2D keypoints~\cite{hu2024animate, hu2025animate, mimicmotion} or 3D SMPL meshes~\cite{wang2023disco, xu2024magicanimate, zhu2024champ, shao2024human4dit} as guidance. Keypoint-based methods, such as OpenPose~\cite{cao2019openpose} and DWPose~\cite{yang2023effective}, control motion with sparse skeletons, while SMPL-based methods better capture surface geometry, occlusions, and body contours~\cite{wang2023disco, zhu2024champ, shao2024human4dit}. While SMPL guidance improves realism, it may reduce generalization and introduce shape leakage. Furthermore, image-based diffusion models struggle with temporal consistency, and adding external pose modules reduces inference efficiency.

In contrast, we build our method on video diffusion models to leverage their temporal coherence and generalization while incorporating SMPL priors to improve pose-guided generation.

\noindent {\bf Noise Warping for Video Diffusion.} Noise warping has been proposed to generate temporally consistent videos~\cite{goflow, howdo, videoeqm}. These methods control motion by modifying the noise input, e.g., Go-With-The-Flow~\cite{goflow} warps noise according to optical flow. However, optical-flow-based warping often fails for complex human motions due to the non-rigid and articulated nature of the human body. To address this, we propose articulated 3D noise warping, specifically designed for human motion, enabling motion-consistent and temporally coherent video generation.

%% file: sec/3_method.tex
\section{Methodology}


\noindent We conduct experiments on latent video diffusion models (LVDM)~\cite{cogvideox}. In this section, we first introduce the preliminaries of LVDM, and then introduce how to model human motions in a structured unified space by proposing articulated noise warping, which attaches Gaussian noise to the human body surface for motion-consistent noise sampling. Then, we describe how to fine-tune a video diffusion model with the articulated noise warping for controllable human video synthesis. For faithful motion control, we propose a joint appearance-motion learning strategy and motion consistency supervision as shown in Fig. \ref{fig:framework}.

\subsection{Preliminaries} \label{sec:method_vdm}

\noindent Latent video diffusion models (LVDM) \cite{cogvideox, svd} learns a denoising process to simulate the probability distribution within the latent space. To improve computational efficiency, a video $\bm{{x}_0}$  is transformed from the pixel space into a latent space feature $\bm{z_0} = \mathcal{E}(\bm{x_0})$ with a 3D Variational Autoencoder (VAE) encoder~\cite{cogvideox} $\mathcal{E}$. $\bm{x_0} \in \mathbb{R}^{f \times H \times W \times 3}, \bm{z_0} \in \mathbb{R}^{f \times h \times w \times C}$, where $f$ is the number of frames, $C$ is the number of channels, $H, W$ and $h, w$ are the height and width of input video and latent feature, respectively.

Gaussian noise is iteratively added to $\bm{z_0}$ at various timesteps $t$ as 
$
q(\bm{z_t} \mid \bm{z_{t-1}}) = \mathcal{N}(\bm{z_t}; \sqrt{1 - \beta_t} \bm{z_{t-1}}, \beta_t {I}),
$
where $\beta$ represents a sequence schedule. The denoising process is defined as an iterative Markov chain that progressively denoises the initial Gaussian noise $z_T \sim \mathcal{N}(0, I)$ into a clean latent space $z_0$. The LVDM denoising function is commonly implemented with U-Net or Transformers, which is trained to minimize the loss of mean square error as
\begin{equation}
\label{eq:diff}
\mathcal{L}_{diff} = \mathbb{E}_{\bm{z_t}, \bm{c}, t, \epsilon \sim \mathcal{N}(0, I)} \left[ \| \bm{\epsilon} - \epsilon_\theta(\bm{z_t}; \bm{c}, t) \|_2^2 \right]
\end{equation}

where $\epsilon_\theta$ represents the parameterized diffusion model for predicting noise and $c$ denotes an optional conditional input (\eg image or text). Subsequently, the denoised latent feature is decoded into video frames (appearance) using the VAE Decoder $\mathcal{D}_A$.

\figFramework

\subsection{Articulated Motion-Consistent Noise Sampling} \label{sec:method_sample}


\noindent \textbf{Structured Noise Texture UV Map.} Existing video diffusion models learn human motions from unstructured random noises, ignoring the human body structure and motion consistency across frames, which leads to motion flickering for human video generation. Instead, we propose to model human motions in a structured unified space, \ie, on the texture UV map of a parametric body model, which is shared across different motion frames for consistent structure-aware motion modeling. We randomly sample a Gaussian noise texture map $\bm{\epsilon_{uv}} \in \mathbb{R}^{{U} \times {V} \times C} $ where $C$, ${U}$, ${V}$ are the number of channels, the height and width of the texture UV map. Note that $\bm{\epsilon_{uv}}$is time-independent and does not vary with the length of the video sequence, which enables motion-consistent human video generation. 

\noindent \textbf{Motion Representation.} We further extract 3D human motions \cite{easymocap} from a sequence of time-varying parametric posed body meshes (\eg SMPL) obtained from training videos. We render these meshes and record the UV coordinate maps \cite{densepose} in image space as motion representations $m$ for each time step $t$.
 
\noindent \textbf{Motion-conditioned Articulated Noise Warping.} We warp the noise texture map $\bm{\epsilon_{uv}}$ from UV space to image space via a geometric warping operation $\mathcal{W}$ that is pre-defined by the parametric body template (e.g., SMPL), which yields a warped noise tensor $\bm{\epsilon} = \mathcal{W}(\bm{m}, \bm{\epsilon_{uv}})$. $\bm{\epsilon} \in \mathbb{R}^{f \times h \times w \times C }$, same shape with $\bm{z_0}$. Note that besides human body, we also add Gaussian noise to the background to model diverse backgrounds in videos.

\subsection{Model Fine-Tuning} \label{sec:method_ft}


\noindent Our video diffusion model is conditioned on the warped motion noise. In particular, we fine-tune the image-to-video (I2V) variant of a latent video diffusion model CogVideoX \cite{cogvideox}. During training, \nickname uses the warped noise instead of regular Gaussian noise.

Different from existing noise warping methods (e.g., Go-with-the-Flow \cite{goflow, videoeqm}) that train the video diffusion model with normal finetuning objective, \ie, the mean squared loss between denoised samples and samples with noise added, we notice that the standard training objective fails to generate temporally consistent videos with faithful motion controls. Because of articulated body structures and non-rigid movements, human motions are far more complex than the rigid object motions in \cite{goflow}. Moreover, as the human visual system is highly sensitive to imperfections in human motion, human video generation typically demands high precision in motion control. To address these challenges, we further introduce several motion learning strategies described below.

\noindent\textbf{Motion Noise Degradation.} We use the strategy of {noise degradation} \cite{goflow} to control the {strength} of motion conditioning at inference time. After calculating the clean warped noise, we then degrade it by a random degradation level $\gamma \in [0,1]$, by first sampling uncorrelated Gaussian noise $\zeta \sim \mathcal{N}(0,1)$ and modifying the warped noise via $\mathcal{T}$:
\begin{equation}
\label{eq:deg}
\mathcal{T}(\bm{\epsilon}) = \frac{(1-\gamma)\bm{\epsilon} + \zeta\gamma}{\sqrt{(1-\gamma)^2 + \gamma^2}}    
\end{equation}

As degradation level $\gamma \to 1$, $\bm{\epsilon}$ approaches an uncorrelated Gaussian, and as $\gamma \to 0$, $\bm{\epsilon}$ approaches clean warped noise. In inference, the user can control how strictly the resulting video should adhere to the input flow. Please see the appendix. for a qualitative depiction of the effect of $\gamma$.


\noindent\textbf{Joint Appearance-Motion Learning (JAML).} Despite tremendous recent progress, generative video models still struggle to capture the dynamics and physics of human motions. This limitation arises from the conventional pixel reconstruction objective, which biases models toward appearance fidelity at the expense of motion coherence \cite{surmo,videojam}. To further enforce faithful motion control, we propose a joint appearance-motion learning strategy that encourages the video diffusion model to learn physical motions by jointly predicting appearance and motion from motion-conditioned noises during training. In addition to the latent diffusion supervision, we propose a motion decoder $\mathcal{D_M}$ to predict motions $\bm{m'}$ from denoised latent $\bm{m'} = \mathcal{M_D}(\bm{z'_{t-1}})$, and the prediction loss optimization loss is defined as:
\begin{equation}
\label{eq:md}
\begin{split}
    \mathcal{L}_{md} = \|\bm{m} - \bm{m'} \|_2 = \| \bm{m} - \mathcal{M_D} \circ \epsilon_\theta(\bm{z_t}; \bm{c}, t)  \|_2
\end{split}
\end{equation}

where $\bm{z_t}$ is the noised sample at time step $t$. In practice, we predict motions from the predicted denoised latent $\bm{z'_0}$.

Note that similar joint appearance-motion learning strategies have been proposed in related papers, such as SurMo \cite{surmo} and VideoJAM \cite{videojam}. However, ours is distinguished from these works: 1) Surmo is designed for 3D human-specific reconstruction. 2) VideoJAM models motions by optical flow and requires slow progressive inner-guidance for video generation during inference. 

 

\noindent \textbf{Geometric Motion Consistency Learning (GMCL).} The standard training objective of LVDM mainly supervises the diffusion loss that focuses on the pixel reconstruction quality while failing to capture the motion consistency between consecutive frames, which causes motion flickering, especially for complex human motions. To tackle this issue, we propose geometric motion consistency supervision to encourage the LVDM to learn motion consistency during diffusion. 

However, the 3D VAE of the CogVideoX model does not faithfully preserve the spatial information after video compression in the latent space, \ie  
the denoised latents do not exhibit strict spatial alignment with the input motion. In contrast, the warped noise $\bm{\epsilon}$ and predicted added noise $\bm{\epsilon'}$ are motion-conditioned and spatially aligned with the input motion.

We first apply degradation reversal to the predicted noise $\bm{\epsilon'}$ as opposite to noise degradation, 

\begin{equation}
\label{eq:reverse_deg}
\mathcal{T}^{-1}(\bm{\epsilon'}) = \frac {\sqrt{(1-\gamma)^2 + \gamma^2}\bm{\epsilon'} - \zeta\gamma}{(1-\gamma)}
\end{equation}

$\bm{\epsilon'}$ is further unwarped into the UV space via inverse warping $\mathcal{W}^{-1}$ and then fused in UV space via average pooling $\mathcal{F}$, which yields $\bm{\epsilon'_{uv}} = \mathcal{F} \circ \mathcal{W}^{-1} \circ \mathcal{T}^{-1} (\bm{\epsilon'})$. The motion consistency loss is calculated as:

\begin{equation}
\label{eq:mc}
\mathcal{L}_{mc} = \|\bm{\epsilon_{uv}} - \bm{\epsilon'_{uv}} \|_2 = \| \bm{\epsilon_{uv}} - \mathcal{F} \circ \mathcal{W}^{-1} \circ \mathcal{T}^{-1} \circ \epsilon_\theta(\bm{z_t}; \bm{c}, t) \|_2
\end{equation}

\noindent where $\mathcal{W}^{-1}$, $\mathcal{T}^{-1}$, $\mathcal{F}$ are all differentiable. Note that we apply mask for $\bm{\epsilon_{uv}}$ and $\bm{\epsilon'_{uv}}$ in UV space when we calculate the loss.

\noindent\textbf{Optimization.} \nickname{} is trained end-to-end to optimize VLDM $\epsilon_\theta$ and motion decoder $\mathcal{D_M}$, whereas the 3D VAE $\mathcal{E}$ and video decoder $\mathcal{D_A}$ are fixed. The total loss is defined as: 
\begin{equation}
\label{eq:totalloss}
\mathcal{L} = \lambda_{diff} \mathcal{L}_{diff} + \lambda_{mc} \mathcal{L}_{mc} + \lambda_{md} \mathcal{L}_{md}
\end{equation}

where the weights $\lambda_{diff}$, $\lambda_{mc}$, and $\lambda_{md} $ are set experimentally. Refer to the appendix. for more details.


\subsection{Inference with Articulated Noise Sampling} \label{sec:method_infer}

We apply \nickname for image to video (I2V) generation, \ie, generating an animation sequence from the input image. In inference, given a single image of a person and a driven motion sequence, we generate motion-conditioned warped noise to guide the motion of the output video. The warped noise is used to initialize the diffusion process of \nickname, and a deterministic sampling process is employed for latent denoising. Compared with existing human video generation methods (e.g., DreamPose \cite{dreampose}) which trains a specific pose module to process poses, our method can fully utilize the video prior of the base model for {intrinsic control}, as it requires no adapters or modifications to the diffusion pipeline, while maintaining the same memory usage and runtime as the base model.

Our method is agnostic to diffusion model design (e.g., CogVideoX-I2V and Wan2.1-I2V \cite{wan2025}). Though the base model CogVideoX-5B I2V ~\cite{cogvideox} is constrained to generate 49 frames during inference, thanks to the novel motion-consistent noise sampling strategy, \nickname{} is capable of generating \textbf{longer video sequences} while maintaining temporal consistency. Refer to more details in the supplementary video.

\subsection{Implementation Details} \label{sec:method_implementation}

We fine-tune the base video diffusion model, the I2V of CogVideoX-5B \cite{cogvideox} for image-to-video tasks, which generates $480\times720$px videos. We follow the same setup of CogVideoX-5B for video encoding, \ie, given a video $\bm{x_0}$ of size $(4f + 1) \times 8h \times 8w \times 3$, the latent embedding $\bm{z_0}$ is encoded by the 3D VAE $\mathcal{E}$, where $\bm{z_0} \in \mathcal{R}^{(f+1)\times h \times w \times C}$. CogVideoX-5B generates 49 frames for each video, \ie, $f=12$, $h=60$, $w=90$, $C=16$. 

We also fine-tune on Wan2.1-I2V \cite{wan2025} to show our method is model-agnostic. Unless otherwise specified, we adopt the variant trained on CogVideoX-5B same as GWTF \cite{goflow} for fair comparisons.

For noise warping, because the diffusion model works on latent embeddings, we estimate the 3D pose in image space via \cite{easymocap} and warp noise in image space, and then downsample that noise into latent space, similar to the video downscaling of CogVideoX by a factor of $8 \times 8$ spatially and $4$ temporally. We use nearest-neighbor interpolation along both the temporal axis and the two spatial axes.

During training, we randomly sample $\gamma \in [0,1]$, and fix it to 0.1 at inference. More details can be found in the appendix.



%% file: sec/4_results.tex

\section{Experiments}
\label{sec:exp}

\figCmpMotionMainAndTabMain

\noindent \textbf{Data Processing}.
We evaluated our method on the UBC Fashion \cite{ubcfashion} dataset, which contains 600 monocular human videos featuring natural fashion motions, with subjects wearing various styles, including dresses. We first estimate 3D SMPL parameters from the monocular videos using \cite{easymocap}, and then crop and scale each video to $480 \times 720$ px (the standard input format for CogVideoX) for training.

\noindent \textbf{Evaluation Metrics.} We evaluate our method using standard metrics widely adopted in prior work. For single-frame image quality, we report the L1 error, Structural Similarity Index (SSIM)~\cite{wang2004ssim}, Learned Perceptual Image Patch Similarity (LPIPS)~\cite{zhang2018lpips}, and Fréchet Inception Distance (FID) \cite{Heusel2017GANsTB}. 
Video fidelity is assessed using Fréchet Inception Distance with Fréchet Video Distance (FID-VID)~\cite{fvd2018} and Fréchet Video Distance (FVD)~\cite{unterthiner2019fvd}.

\noindent \textbf{Baselines}. We compare our approach with a range of state-of-the-art baselines. MRAA \cite{mraa} represents a leading GAN-based animation method that predicts optical flow from the driving sequence to warp the source image, followed by GAN-based inpainting to handle occluded regions. We also include Thin-Plate Spline Motion Model (TPSMM), the diffusion-based pose transfer method PIDM \cite{pidm}, and the bidirectionally deformable motion modulation approach BDMM. DreamPose \cite{dreampose} and Animate Anyone \cite{hu2024animate} leverage the Stable Diffusion prior \cite{sd} for human animation  generation. All these methods are trained on the UBC Fashion dataset for fair evaluation. UniAnimate \cite{wang2025unianimate} adopt the Wan2.1 video base model for controllable human video synthesis.

In addition, we compare against generic video diffusion models, including Stable Video Diffusion (SVD-I2V) \cite{svd}, our base model CogVideoX-I2V, and the motion-controllable video diffusion framework Go-With-The-Flow (GWTF) \cite{goflow}.

\subsection{Quantitative Analysis}


We test all models on the UBC Fashion test set, which consists of 100 fashion videos. For fair comparison, we scale our results to a 256 px resolution, the same as DreamPose \cite{dreampose}. For each video, we use the generated 49 frames for testing. The quantitative results are listed in Tab. \ref{tab:main}, where the metrics of MRAA, TPSMM, PIDM, SVD-I2V, and DreamPose are reported in \cite{dreampose,hu2024animate}. The results indicate that our method outperforms existing baselines by a large margin on both image metrics (L1, SSIM, LPIPS, FID) and video metrics (FVD, FID-VID). This demonstrates that our method can generate not only high-quality images but also temporally consistent human videos. It should be mentioned that our model and Go-With-The-Flow (GWTF) are both built upon CogVideoX base model, whereas our model significantly outperforms the others for human video generation.

\subsection{Qualitative Analysis}

We qualitatively compare our method to DreamPose \cite{dreampose}, TPSMM \cite{tpsmm}, BDMM \cite{bdmm}, SVD-I2V \cite{svd}, CogVideoX-5b-I2V \cite{cogvideox}, and GWTF \cite{goflow}, as shown in Fig. \ref{fig:cmp_main}. For DreamPose, BDMM, and SVD-I2V, note that the person identity, fabric folds, and fine patterns are lost in new poses, whereas our method can accurately retain those details even for complex texture styles. 

Since GWTF and our model are both built upon CogVideoX and use the same video VAE model, they are able to preserve the appearance details of the reference image. However, neither CogVideoX nor GWTF supports faithful fine-grained human motion control, \ie, the rendered images are not fully aligned with the input poses. In contrast, thanks to articulated 3D noise sampling, joint appearance-motion learning, and our geometric motion consistency learning strategy, our method can generate high-quality human motion videos. Refer to the appendix. for more details.

\figSupCmpAAMotionFlow

We also compare our method with Animate Anyone \cite{hu2024animate} and Wan2.1 based UniAnimate \cite{wang2025unianimate} for fashion video generation, and the qualitative results are shown in Fig. \ref{fig:sup_cmpaa}, which suggest that our method can generate high-quality clothing details (\eg, \cnum{1}, \cnum{3}), and high-quality face datails (\cnum{2}), whereas Animate Anyone cannot achieve the same level of realism. Note that Animate Anyone and our method were both fine-tuned on fashion videos, whereas the original UniAnimate was fine-tuned on generic videos.

\noindent \textbf{3D Articulated Noise Warping vs. 2D Optical Flow based Noise Warping.} We further analyze two different noise warping methods: 2D optical flow based GWTF \cite{goflow}, and our 3D articulated noise warping. The qualitative comparisons are shown in Fig. \ref{fig:cmp_goflow}, which indicates that our 3D noise warping is more suitable for human rendering, \ie, it enables more precise motion control, and generates more consistent rendering results even for complex clothing fashion style textures. More comparisons can be found in the supplementary video.


\figGeneration

\subsection{Generalization on Out-of-domain Data}
 
Though our model was fine-tuned on studio fashion videos, it generalizes well on out-of-domain data. We evaluate the generalization capability on DeepFashion dataset \cite{Liu2016DeepFashion} and in-the-wild publicly available videos.

\noindent \textbf{Human Animation on DeepFashion.} Our method generalizes well to unseen out-of-domain data, producing temporally consistent videos from a single image of humans in diverse clothing styles, as shown in Fig. \ref{fig:sup_df}. More results can be found in the supplementary video.

\noindent \textbf{Human Animation on In-the-Wild Videos.} Our method generalizes well to unseen in-the-wild videos collected from publicly available online sources with diverse backgrounds, producing temporally consistent videos from a single image, as shown in Fig. \ref{fig:sup_sm}. More results can be found in the supplementary video.


%

\subsection{Ablation Study}


\figAbMd
\noindent \textbf{Joint Appearance-Motion Learning (JAML).} We analyze the effectiveness of the JAML mechanism (Sec. \ref{sec:method_ft}) qualitatively and quantitatively. The qualitative comparisons are shown in Fig. \ref{fig:ab_md}, which suggests that the joint learning strategy improves motion coherence for faithful motion control, while the variant that was trained with a standard objective  (w/o JAML) failed to generate accurate animation videos. The quantitative results listed in Tab. \ref{tab:ab} also indicates that the joint learning strategy can improve the quantitative performance.  

\figAbMc

\noindent \textbf{Geometric Motion Consistency Learning (GMCL).} 
We also analyze the effectiveness of our proposed Geometric Motion Consistency Learning strategy. The qualitative comparisons are shown in Fig. \ref{fig:ab_mc}, where we visualize two different frames for consistency comparison. Fig. \ref{fig:ab_mc} shows that the proposed GMCL can significantly improve appearance consistency. For example, the face identity and hand details are better preserved, \eg, \cnum{2} vs. \cnum{1}, and \cnum{4} vs. \cnum{3}. 
 
\tabAb
Tab. \ref{tab:ab} summarizes the quantitative results, which suggest that the quantitative results are significantly improved by a big margin with the GMCL strategy.


%% file: sec/discussion.tex
\section{Discussion and Conclusion}





\nickname{} is a \textbf{model-agnostic} framework that extends base video diffusion models for controllable and longer human video generation without compromising efficiency. 


\figWanCog

$\bullet$ \textbf{Model Agnostic}. We fine-tune on different base video diffusion models including CogVideox-I2V-5B \cite{cogvideox} and  Wan2.1-I2V-14B \cite{wan2025} to show our method is model-agnostic, and the generation results are shown in Fig. \ref{fig:wancog}. The model trained on Wan2.1-I2V-14B generates higher quality appearance details than that of CogVideox-I2V-5B. 

$\bullet$ \textbf{Efficiency}. \nickname{} shares the same architecture as the base video diffusion model, and
hence uses exactly the same amount of memory and runtime as the base model for both  CogVideoX \cite{cogvideox} and Wan2.1 \cite{wan2025}. 

$\bullet$ \textbf{Generate Longer Video Sequences.} Our method can extend the original framework to generate longer video sequences, e.g., from 49 to 81 frames, using motion-consistent noise sampling without any changes to the architecture. More results can be found in the supplementary video.

\noindent \textbf{Conclusion.} We presented \nickname{}, a novel framework for controllable human video generation that combines articulated motion-consistent noise sampling, joint appearance-motion learning, and geometric motion consistency training. Our approach enables high-fidelity motion-consistent human video synthesis without requiring additional motion modules for intrinsic motion control at inference. Extensive experiments demonstrate that \nickname{} outperforms prior methods, providing a scalable and effective solution for image-to-video generation with precise motion control.


%% file: pack/tables_sup.tex
\newcommand{\tabSupAbDeg}{
\begin{table}[h]
\centering
\caption{Ablation study of noise degradation. $\gamma$ measures how strictly the resulting video adheres to the motion input.}

\begin{tabular}{lcc}
\hline
& LPIPS $\downarrow$ & SSIM $\uparrow$ \\
\hline
$\gamma=0$ & .0857 & .858 \\
$\gamma=0.1$ & \textbf{.0821} & \textbf{.860} \\
$\gamma=0.2$ & .0859 & .860 \\
$\gamma=0.3$ & .0898 & .858 \\
$\gamma=0.4$ & .1088 & .841 \\
$\gamma=0.5$ & .1184 & .834 \\
$\gamma=0.6$ & .1159 & .836 \\
$\gamma=0.7$ & .1031 & .852 \\
$\gamma=0.8$ & .1134 & .841 \\
$\gamma=0.9$ & .1055 & .845 \\
$\gamma=1.0$ & .1209 & .832 \\
\hline
\end{tabular}
\label{tab:sup_ab_deg}
\end{table}
}

%% file: sec/sup_opt.tex

\appendix

\section{Appendix: Implementation Details} 


\noindent \textbf{Network Architecture.} 
We fine-tune on different base video diffusion models for image to video generation, including CogVideox-I2V-5B \cite{cogvideox} and  Wan2.1-I2V-14B \cite{wan2025}. For motion decoder, to accelerate training, we use a CNN-based architecture (Pix2PixHD \cite{pix2pixhd}) with 2 Decoder blocks of [ReLU, ConvTranspose2d, BatchNorm]. 

\noindent \textbf{Noise UV Texture Map.} The size of the noise texture map is $128\times128\times16$.  
 

\noindent \textbf{Optimization.} We used 8 NVIDIA H100 80GB GPUs over the course of 32 GPU days, for 26,000 iterations using a rank-2048 LoRA \cite{Hu2021LoRALA} with a learning rate of 1E-4 and a batch size of 6. 

We set the weights $\lambda_{diff}=1.0$, $\lambda_{mc}=0.5$, and $\lambda_{md}=0.5$  experimentally during training.

%% file: sec/sup_exp_org.tex

\section{Additional Ablation Study}

\figSupAbDeg
\tabSupAbDeg

\noindent \textbf{Noise Degradation.} We use the noise degradation strategy~\cite{goflow} to control the strength of motion conditioning at inference time. We analyze the effect of the noise degradation parameter $\gamma$ for human video generation on a subset of the UBC Fashion dataset. Table~\ref{tab:sup_ab_deg} summarizes the quantitative results, suggesting that our method achieves the best performance on both LPIPS and SSIM when $\gamma = 0.1$.

We also show qualitative results in Fig.~\ref{fig:sup_ab_deg}. The generated videos strictly follow the motion signals when $\gamma = 0$, but the video quality is degraded. Therefore, we set $\gamma = 0.1$ in our experiments to achieve a balance between motion control and video quality.




\figVAE
\section{Limitations}

Our model relies on the video VAE of the underlying video diffusion backbone to encode input videos, and the overall generation quality is therefore constrained by the VAE’s reconstruction fidelity. In particular, the video VAE used in CogVideoX tends to lose fine-grained details—such as facial features and hand structures—during the encoding process. Consequently, models built upon CogVideoX struggle to synthesize high-quality facial and hand details, as illustrated in Fig. \ref{fig:vae}. A promising direction for future work is to explore stronger video diffusion backbones equipped with more expressive VAEs, or to jointly fine-tune the VAE together with the diffusion model, following prior work such as \cite{dreampose}.